\documentclass[sigconf]{acmart}

\AtBeginDocument{%
  }
\usepackage{enumerate}
\usepackage{booktabs}
\usepackage{algorithm}
\usepackage[noend]{algpseudocode}
\usepackage{bm}
\usepackage{subcaption}

\usepackage{balance}
\usepackage{multirow}
\usepackage{multirow}
\usepackage{float}
\usepackage{url}
\usepackage{footnote}
\usepackage{mathrsfs}
\usepackage{amsthm,amsmath}
\usepackage{caption}

\usepackage{color,xcolor}
\definecolor{shapecolor}{rgb}{0.0,0.5,0.0}
\usepackage{multirow}
\usepackage{pifont}

\copyrightyear{2025}
\acmYear{2025}
\setcopyright{acmlicensed}\acmConference[MM '25]{Proceedings of the 33rd ACM International Conference on Multimedia}{October 27--31, 2025}{Dublin, Ireland}
\acmBooktitle{Proceedings of the 33rd ACM International Conference on Multimedia (MM '25), October 27--31, 2025, Dublin, Ireland}
\acmDOI{10.1145/3746027.3755131}
\acmISBN{979-8-4007-2035-2/2025/10}

\acmSubmissionID{2222}

\begin{document}
\title{UIS-Mamba: Exploring Mamba for Underwater Instance Segmentation via Dynamic Tree Scan and Hidden State Weaken}

\author{Runmin Cong}
\authornote{Also with the Key Laboratory of Machine Intelligence and System Control, Ministry of
Education, Jinan, Shandong, China.}
\orcid{0000-0003-0972-4008}
\affiliation{%
  \institution{School of Control Science and Engineering, Shandong University}
  \city{Jinan}
  \state{Shandong}
  \country{China}
}
\email{rmcong@sdu.edu.cn}

\author{Zongji Yu}
\authornotemark[1]
\orcid{0009-0007-4146-6535}
\affiliation{%
  \institution{School of Control Science and Engineering, Shandong University}
  \city{Jinan}
  \state{Shandong}
  \country{China}
}
\email{zjyu@mail.sdu.edu.cn}

\author{Hao Fang}
\authornotemark[1]
\authornote{Corresponding author.}
\orcid{0000-0002-8846-8294}
\affiliation{%
  \institution{School of Control Science and Engineering, Shandong University}
  \city{Jinan}
  \state{Shandong}
  \country{China}
}
\email{fanghaook@mail.sdu.edu.cn}

\author{Haoyan Sun}
\authornotemark[1]
\orcid{0009-0001-9373-9685}
\affiliation{%
  \institution{School of Control Science and Engineering, Shandong University}
  \city{Jinan}
  \state{Shandong}
  \country{China}
}
\email{heysun@mail.sdu.edu.cn}

\author{Sam Kwong}
\orcid{0000-0001-7484-7261}
\affiliation{%
  \institution{School of Data Science, Lingnan University}
  \city{Hong Kong}
  \country{China}
}
\email{samkwong@ln.edu.hk}

\begin{abstract}
Underwater Instance Segmentation (UIS) tasks are crucial for underwater complex scene detection. 
Mamba, as an emerging state space model with inherently linear complexity and global receptive fields, is highly suitable for processing image segmentation tasks with long sequence features.
However, due to the particularity of underwater scenes, there are many challenges in applying Mamba to UIS. The existing fixed-patch scanning mechanism cannot maintain the internal continuity of scanned instances in the presence of severely underwater color distortion and blurred instance boundaries, and the hidden state of the complex underwater background can also inhibit the understanding of instance objects.
In this work, we propose the first Mamba-based underwater instance segmentation model UIS-Mamba, and design two innovative modules, Dynamic Tree Scan (DTS) and Hidden State Weaken (HSW), to migrate Mamba to the underwater task. 
DTS module maintains the continuity of the internal features of the instance objects by allowing the patches to dynamically offset and scale, thereby guiding the minimum spanning tree and providing dynamic local receptive fields.
HSW module suppresses the interference of complex backgrounds and effectively focuses the information flow of state propagation to the instances themselves through the Ncut-based hidden state weakening mechanism. Experimental results show that UIS-Mamba achieves state-of-the-art performance on both UIIS and USIS10K datasets, while maintaining a low number of parameters and computational complexity. Code is available at https://github.com/Maricalce/UIS-Mamba.

\end{abstract}



\maketitle
\section{Introduction}
To address the growing demand for underwater exploration, the deep learning vision community has increasingly focused on developing underwater visual tasks~\cite{zhao2020research,fan2023advances}. Underwater Instance Segmentation (UIS) ~\cite{lian2023watermask} and Underwater Salient Instance Segmentation (USIS)~\cite{wang2024underwater} are emerging tasks: UIS aims to segment all instance objects in underwater scenes with category distinction, while USIS specifically targets visually salient objects by differentiating prominent instances. These tasks are fundamentally critical for underwater image enhancement~\cite{cong2023pugan,zhou2025net,jiang2022underwater}, oceanic resource exploration~\cite{jin2024underwater}, human-robot interaction~\cite{pan2025semantic,zhang2024learning,guan2024structural}, and underwater video segmentation~\cite{fang2024learning,fang2024unified,fang2025decoupled}. 
However, existing deep learning models for UIS and USIS remain at a comparatively nascent stage, as visual comprehension of underwater scenes remains a challenging task~\cite{akkaynak2017space}. For example, underwater images inherently suffer from quality degradation due to wavelength-dependent attenuation and scattering effects~\cite{kaneko2023marine}. 
The unique visual content of underwater scenes, characterized by distinctive marine species, water-induced texture patterns, and abundant optical artifacts, differs fundamentally from terrestrial imagery~\cite{wang2024underwater}. Their distinctive instance contours and object features impose significant challenges for model learning.

\begin{figure}[t]
  \centering 
  \includegraphics[width=\linewidth]{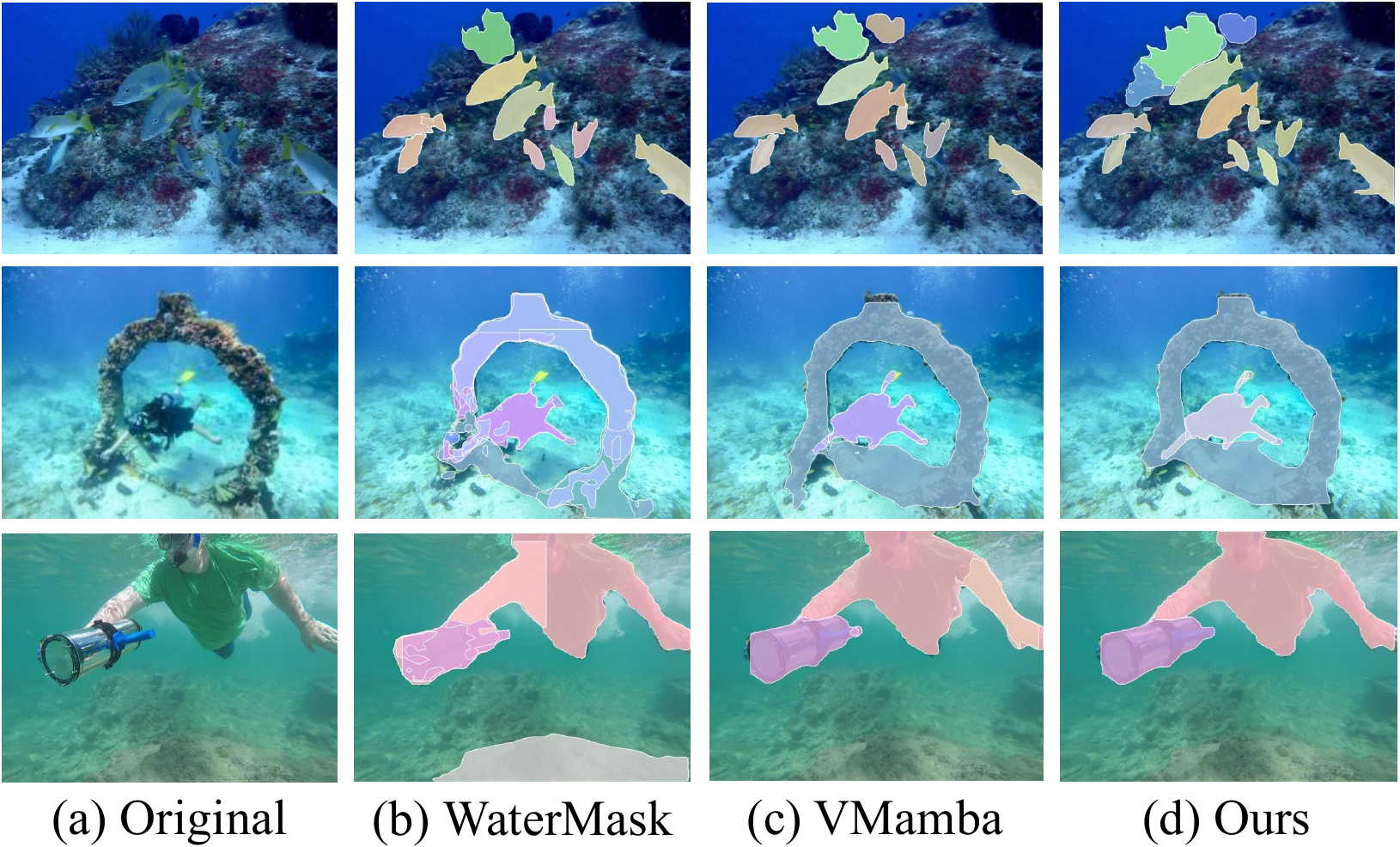}
  \caption{A simple comparison of WaterMask, VMamba and UIS-Mamba on UIIS and USIS10K datasets. The first, second, third and fourth  columns indicate the original image, results of WaterMask, VMamba and our UIS-Mamba, respectively.}
  \label{fig:intro}
\end{figure}

Based on the powerful representation ability of neural networks, many methods based on convolutional neural networks (CNN)~\cite{he2016deep,krizhevsky2012imagenet} have been proposed for instance segmentation tasks. Mask-RCNN~\cite{he2017mask} dominates the instance segmentation task by virtue of lower number of parameters and higher accuracy. WaterMask~\cite{lian2023watermask} migrates the Mask-RCNN to the underwater task by improving the detection head and segmentation head to achieve finer segmentation of underwater targets. Although CNN-based methods have achieved remarkable results, the restricted receptive field limits their ability to capture global dependencies. To better capture global information, methods based on Transformer~\cite{vaswani2017attention,liu2021swin} have been introduced. USIS-SAM~\cite{lian2024diving} applies SAM~\cite{SAM_2023_ICCV} to underwater salient instance segmentation by designing underwater adaptive ViT encoder and salient feature prompter generator. Although the performance has been significantly improved, Transformer architecture also brings a huge number of parameters and computational complexity that cannot be ignored.
Mamba~\cite{gu2023mamba,gu2021efficiently} achieves a balance between efficiency and performance in visual tasks~\cite{liu2024vmamba,vim,guo2024mambair,wan2024sigma} through linear complexity, selective state updates, and hardware optimization. Its core advantage lies in flexible modeling of long sequences and effective utilization of computing resources, which is highly suitable for image segmentation tasks with long sequence features.
To verify the performance of Mamba in underwater tasks, we replaced the backbone of WaterMask~\cite{lian2023watermask} from ResNet~\cite{he2016deep} to VMamba~\cite{liu2024vmamba} to simply introduce Mamba into UIS. As shown in Fig.~\ref{fig:intro} (a,b), although VMamba's segmentation results are slightly better than WaterMask, the potential of Mamba seems to have not been fully exploited. 

We believe that the uniqueness of underwater scenes and the scanning mechanism of existing vision Mamba are potential reasons. There are significant peculiarities of underwater scenes, such as light fading and color distortion of underwater images caused by phenomena such as channel degradation. For underwater instance objects of the same category, they may exhibit vastly different features in different underwater scenes. However, both early vision mamba~\cite{liu2024vmamba,vim} and subsequent work to improve scanning order~\cite{huang2024localmamba,yang2024plainmamba} are fixed-patch scanning mechanisms that cannot maintain the internal continuity of scanned instances in the presence of underwater color distortion and blurred instance boundaries. In addition, Mamba's sequential scanning mechanism treats various underwater environments equally and performs hidden state updating, so the hidden state of complex underwater backgrounds can also hinder the understanding of instance objects. Therefore, introducing advanced state space models into underwater tasks presents significant challenges. Based on this, this paper, inspired by deformable convolution~\cite{dai2017deformable, zhu2019deformable}, tries to design an algorithmic logic for dynamic patch scanning from a new perspective, endowing the state space model with a dynamic local sense field. By distinguishing foreground and background, we suppress the hidden state features of the background, thereby better capturing the characteristics of instance objects.

Based on the above theory, this paper proposes the first Mamba-based underwater instance segmentation model UIS-Mamba, which has dynamic local receptive fields in the scanning phase and can adapt Mamba to the comprehension of complex underwater scenes through hidden state suppression. UIS-Mamba has two key modules: Dynamic Tree Scan (DTS) and Hidden State Weaken (HSW). 
DTS module initially predicts the position of instance objects and the contour features of specific parts during the scanning phase, and establishes a four-connected tree~\cite{xiao2024grootvl} for patches, allowing the patches of each node to dynamically offset and scale according to the local information. This eliminates the feature loss caused by the fixed division of patches, allowing the same part of the instance object to be intact in the same patch, improving model's understanding ability, and guiding the construction of the minimum spanning tree~\cite{yang2014stereo}; HSW module, on the basis of the minimum spanning tree, labels the background patch by Ncut algorithm~\cite{wang2022tokencut,boruvka1926jistem}, and prune the background patch in the hidden state update operation. This weakens the influence of background patches during the update process, effectively avoiding interference from complex underwater environments on the model's ability to capture correct instance objects. By assembling all the contributions into UIS-Mamba and replacing the backbone of WaterMask~\cite{lian2023watermask}, we significantly improve the performance of Mamba in underwater scenes, as shown in Fig.~\ref{fig:intro} (c). Our model achieves state-of-the-art (SOTA) performance on both the underwater instance segmentation dataset UIIS and the underwater salient instance segmentation dataset USIS10K, while maintaining low parameter and computational complexity.

The contributions can be summarized as follows:

\begin{enumerate}[0]
 \item[$\bullet$]We reveal the limitations of existing vision Mamba in underwater scenes and propose first Mamba-based underwater instance segmentation model UIS-Mamba, which improves the vision Mamba's ability to understand the features of underwater instance objects and provides new insights for Mamba's migration to underwater tasks.

 \item[$\bullet$]We design Dynamic Tree Scan (DTS) module and Hidden State Weaken (HSW) module to introduce vision Mamba into underwater scenes. The DTS module maintains scanning continuity of the internal features of the instance objects by allowing the patches to dynamically offset and scale, and the HSW module weakens the interference of the complex underwater background on the hidden state update.

 \item[$\bullet$]Experiments on the UIIS dataset and USIS10K dataset regarding instance segmentation and salient instance segmentation show that our proposed UIS-Mamba achieves state-of-the-art segmentation performance and maintains a low number of parameters and computational complexity.

\end{enumerate}

\section{Related Work}

\subsection{State Space Models}
State space models (SSMs) have emerged as a new class of models in the deep learning paradigm, showing great potential for sequence transformation. These methods have attracted significant attention due to their linear scalability with sequence length. An early approach, LSSL~\cite{gu2021combining}, drew inspiration from continuous state-space models in control systems and attempted to address the problem of remote dependence by combining it with HIPPO~\cite{fu2022hungry} initialization. S4 proposed to normalize the parameters into diagonal matrices, prompting a subsequent series of structured SSMs~\cite{gu2021efficiently} research. Recently, the selective state-space model ~\cite{gu2023mamba}, known as Mamba, has struck a balance between effectiveness and efficiency by designing input-dependent parameter initialization strategies, and it has become a strong competitor to Transformer and CNN structures.

In addition to demonstrating superior results in sequence modeling, Mamba has been extensively introduced into the vision domain. VMamba~\cite{liu2024vmamba} and Vision Mamba~\cite{vim} propose to traverse the spatial domain and convert any non-causal visual image into an ordered sequence of patches, introducing Mamba into image classification task for the first time and demonstrating strong competitiveness. And LocalMamba~\cite{huang2024localmamba} strategically preserves local detail information by local scanning. PlainMamba~\cite{yang2024plainmamba} introduces a continuous scanning strategy aiming to alleviate the semantic discontinuity problem by simply adjusting the propagation direction at the discontinuous locations. Spatial-Mamba~\cite{xiao2025spatialmamba} mitigates the localized receptive fields by incorporating multiscale dilation convolution into Mamba's state update. GrootV~\cite{xiao2024grootvl} targeted scanning of each image by building a connected tree.

A large number of approaches, while mitigating Mamba's modeling deficiencies on 2D noncausal images to varying degrees by changing the scanning order, have failed to improve the root cause of vision Mamba, \emph{i.e.}, semantic cutoff due to patch-fixed chunking in the proximity region. This issue may not have a significant impact on image classification, but it is crucial for pixel-level tasks such as segmentation. To solve this problem, we propose a dynamic tree scanning strategy, which extends Mamba's dynamic local receptive field in disguise by allowing patches that need to be input to the state space model to actively undergo offsetting and scaling, adjusting the semantic information they contain, and thus dividing the same semantic parts of instance objects into a single patch.

\subsection{Underwater Instance Segmentation}

Underwater Instance Segmentation (UIS) is an emerging visual task with two dedicated datasets: the UIIS dataset~\cite{lian2023watermask} and the USIS10K dataset~\cite{lian2024diving}, corresponding to two distinct objectives. UIS focuses on segmenting all instance objects in underwater scenes and distinguishing their categories, while Underwater Salient Instance Segmentation (USIS) aims to segment visually prominent objects~\cite{zhou2021dense,zhou2021edge,jin2025hierarchical} and differentiate individual salient instances.

Current deep learning models for UIS and USIS remain in their infancy. WaterMask~\cite{lian2023watermask} adapts traditional instance segmentation model Mask R-CNN~\cite{he2017mask} to underwater scenes through architectural modifications, while USIS-SAM~\cite{lian2024diving} employs specialized adapters and prompting mechanisms to enhance SAM’s comprehension~\cite{SAM_2023_ICCV} of underwater scenes. However, WaterMask suffers from limited accuracy due to backbone limitations, and USIS-SAM’s high parameter count and task-specific prompting design fail to handle scenarios~\cite{zhang2024towards} with densely populated instances (potentially numbering in the hundreds) in underwater imagery.

To address these challenges in complex and diverse underwater environments, we propose a Mamba-based underwater instance segmentation architecture. It achieves dual compatibility with both general and salient instance segmentation tasks while maintaining low parameter counts and computational complexity. By dynamically mitigating the impact of complex underwater backgrounds, our method prioritizes instance-centric feature learning.

\begin{figure*}[htbp]
	\center{\includegraphics[width=18cm]  {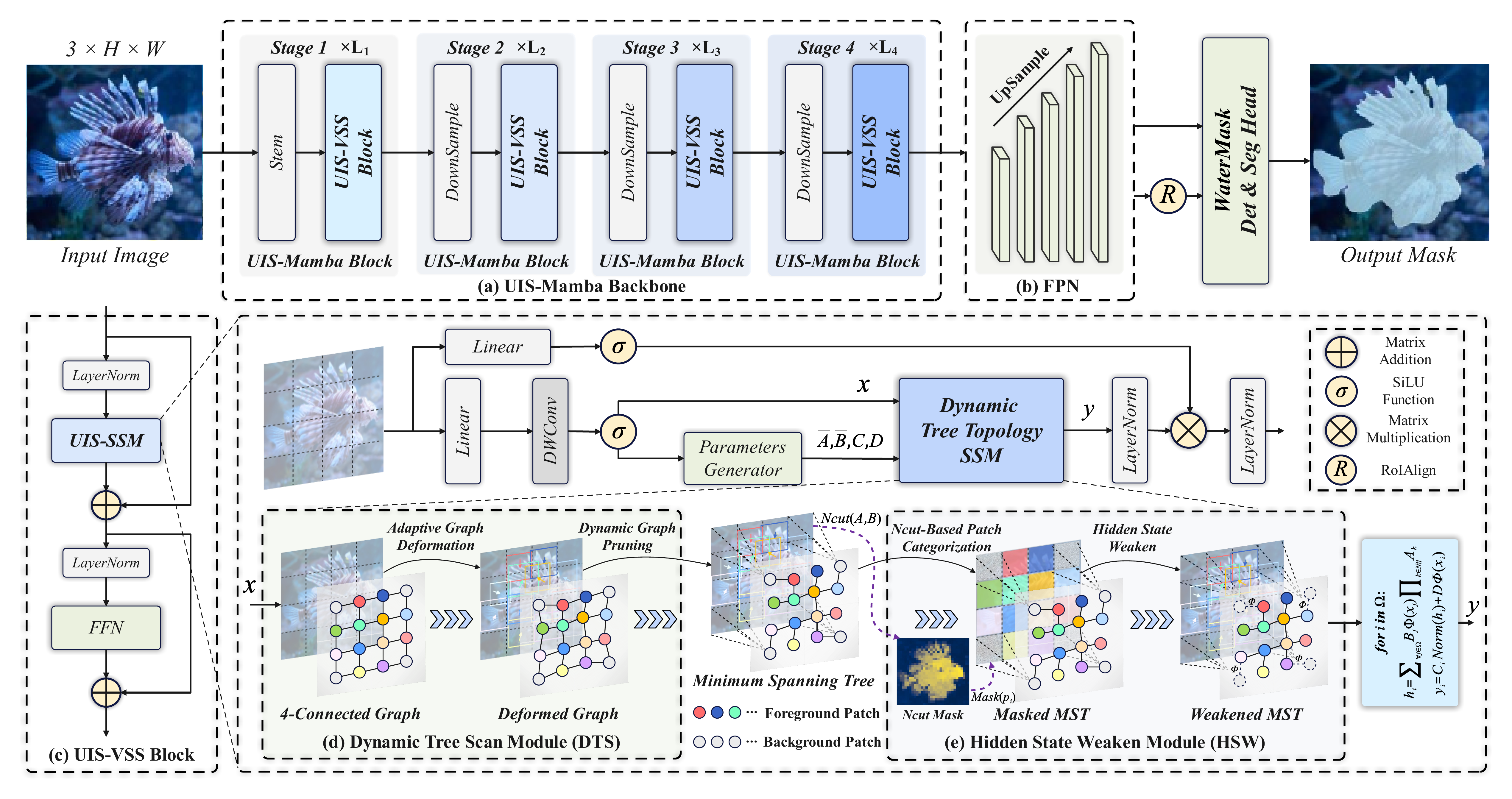}} 
	\caption{Framework of UIS-Mamba. UIS-Mamba include: (a) The UIS-Mamba Backbone; (b) The FPN Module; (c) The UIS-VSS Block; (d) The Dynamic Tree Scan Module (DTS); (e) The Hidden State Weaken Module (HSW). Our major improvements are contained in the Dynamic Tree Topology SSM and use the same detection and segmentation heads as WaterMask.}
    \label{framework}
\end{figure*}

\section{Preliminaries}
State space model (SSM) is commonly used to analyze sequential data and continuous linear time-varying systems~\cite{williams2007linear}, which maps the input $x(t) \in \mathbb{R}^{1 \times D}$ to the output signal $y(t) \in \mathbb{R}^{1 \times N}$ via the state vector $h(t) \in \mathbb{R}^{1 \times N}$, where $t$, $D$ and $N$ denote the step size, the number of signal channels and the state size, respectively. This type of kinetic model can be expressed as the following state and output equations:

\begin{equation}
\begin{aligned}
    \dot{h}(t) &= \mathbf{A}h(t) + \mathbf{B}x(t), \\
    y(t) &= \mathbf{C}h(t) + \mathbf{D}x(t).
\end{aligned}
\end{equation}

To introduce SSM into deep learning frameworks, most approaches choose to discretize the continuous system equations~\cite{gu2023mamba}, using the zero-order holding (ZOH) technique, and convert continuous variables ($\mathbf{A}$, $\mathbf{B}$, $\mathbf{C}$, $\mathbf{D}$) into corresponding discrete parameters\cite{vim,huang2024localmamba,liu2024vmamba} ($\bar{\mathbf{A}}$, $\bar{\mathbf{B}}$, $\bar{\mathbf{C}}$, $\bar{\mathbf{D}}$) over the specified sampling time-scale $\Delta \in \mathbb{R}^{D}$:

\begin{equation}
\begin{aligned}
    \bar{\mathbf{A}} = e^{\Delta \mathbf{A}}, \
    \bar{\mathbf{B}} = (e^{\Delta \mathbf{A}} - \mathbf{I}) \mathbf{A}^{-1} \mathbf{B}, \
    \bar{\mathbf{C}} = \mathbf{C}, \
    \bar{\mathbf{D}} = \mathbf{D}.\
\end{aligned}
\end{equation}

Mamba introduces a dynamic mechanism to selectively filter out input into a sequential state. Specifically, it utilizes linear projection to calculate the parameters $\{\mathbf{B}_i, \mathbf{C}_i\}_{i=1}^L$ for the input sequence $\{\mathbf{x}_i\}_{i=1}^L$, where $\mathbf{x}_i \in \mathbb{R}^{1 \times D}$, directly to improve the context-aware ability. Then the output sequence $\{\mathbf{y}_i\}_{i=1}^L$ is computed with those input-adaptive discretized parameters as follows:
\begin{equation}
\begin{aligned}
    \mathbf{h}_i &= \mathbf{A}_i \mathbf{h}_{i-1} + \mathbf{B}_i \mathbf{x}_i, \\
    \mathbf{y}_i &= \mathbf{C}_i \mathbf{h}_i + \mathbf{D} \mathbf{x}_i,
\end{aligned}
\end{equation}
where $\mathbf{h}_i$ is the hidden state of the state-space model who performs state updates within each discrete step and acts on the output $\mathbf{y}_i$ via the matrix $\mathbf{C}_i$.

\section{UIS-Mamba}
In this section, we introduce our proposed UIS-Mamba, an underwater instance segmentation architecture based on state space modeling. The framework of UIS-Mamba is illustrated in Fig.\ref{framework}. In UIS-Mamba, we design the Dynamic Tree Scan Module (DTS) and Hidden State Weaken Module (HSW) to migrate the vision Mamba to the underwater instance segmentation task.

\subsection{Dynamic Tree Scan Module (DTS)}
Underwater instance objects have complex class features, and the same class instances also have unique part features and local contours. Mamba's existing scanning mechanism with fixed patches is unable to maintain the internal continuity of the scanned instances in the presence of underwater color distortion and instance boundary blurring. 

To address these issues, we propose the Dynamic Tree Scan Module (DTS), which allows patches to be dynamically offset and scaled to maintain the continuity of the internal features of the instance objects through a two-step process of Adaptive Graph Deformation and Dynamic Graph Pruning, and guides the minimum spanning tree by combining spatial and semantic information. Adaptive Graph Deformation dynamically adjusts the feeling field to preserve topology, while Dynamic Graph Pruning eliminates redundant feature connections. This dual mechanism enables context-aware information extraction and feature integration, effectively communicating global-local representation learning for underwater visual recognition.

\noindent \textbf{Adaptive Graph Deformation.} 
We first input the image $x$ that has been processed by downsampling and other modules~\cite{xiao2024grootvl}.
Then, we build a quad-connected graph for the fixed patches, where each patch is essentially a node in the graph. For the set of patches $\mathcal{P} \in \mathbb{R}^{H\times W\times C}$, we pass the dense prediction function $f_p(\cdot)$ for all the patches and multiply them by the weight coefficients with the activation function to generate the adaptive deformation parameters by double parallel projection:

\begin{equation}
\begin{aligned}
    \Delta x, \Delta y &= \text{Tanh}(\mathbf{W}_{\text{offset}} \cdot f_p(\mathbf{p})), \\
    \Delta w, \Delta h &= \text{Softplus}(\mathbf{W}_{\text{scale}} \cdot f_p(\mathbf{p}) + \mathbf{b}),
\end{aligned}
\end{equation}
where the learnable offset parameter $(\Delta x, \Delta y)$ can be adaptively offset according to the local pixel distribution of the feature map to include pixel regions of the same parts as much as possible. The scale factor $(\Delta w, \Delta h)$ can dynamically adjust the patch size to focus on different parts of the various underwater instance objects. The position and range of the deformed patch nodes $\tilde{\mathbf{p}}$ are calculated by bilinear interpolation using the predicted offset and scale:
\begin{equation}
    \tilde{\mathbf{p}} = G(\mathbf{p}\left(p_x + \Delta x, p_y + \Delta y\right); \Delta w, \Delta h)),
\end{equation}
where $\mathbf{p}$ is a patch in the set $\mathcal{P}$, $\tilde{\mathbf{p}}$ is the deformed patch. $p_x$ and $p_y$ are the coordinates of the centers in the x and y directions of $\mathbf{p}$. The bilinear kernel $G(\cdot)$ with a learnable scaling factor preserves the gradient continuity at the edges of the underwater object. Even under severe occlusion or low-contrast conditions, this adaptive deformation can accurately localize the boundaries of the underwater instance objects, yielding a deformed quadratic connectivity graph.

\noindent \textbf{Dynamic Graph Pruning.} For the deformed 4-connected graph $\mathcal{G}$ that has been constructed, in order to overcome the discontinuity of the two-dimensional semantics in the traditional sequential scanning method, and at the same time to fully and adaptively utilize the feature graph information obtained from Adaptive Patch Deformation, we comprehensively improve the calculation method of edge weights by mixing the spatial and semantic information of graph $\mathcal{G}$. Specifically, we add distance measurement to the original patch similarity to calculate the weight $w_{ij}$ of dynamic edge $e_{ij}$ and generate a Minimum Spanning Tree (MST) $\mathcal{T}$:

\begin{equation}
    w_{ij} = \alpha \cdot \|\mathbf{\tilde{p}}^c_{i} - \mathbf{\tilde{p}}^c_{j}\|_2 + (1-\alpha) \cdot \text{Cosine}(\mathbf{\tilde{p}}_i, \mathbf{\tilde{p}}_j),
\end{equation}

\begin{equation}
\mathcal{T}=\mathrm{MST}(\mathcal{G})=\arg\min_{\mathcal{G}}\sum_{e_{ij}\in \mathcal{G}}\text{exp}(w_{ij}),
\end{equation}
where the $\mathbf{\tilde{p}}_i$ and $\mathbf{\tilde{p}}_j$ represent the deformed patch $i$ and patch $j$, $c$ represents the coordinates of the patch. The learnable parameter $\alpha \in [0,1]$ automatically balances spatial adjacency and semantic similarity, particularly crucial for distinguishing overlapping marine organisms with similar appearances. $\mathcal{T}$ generated by Contractive Boruvka algorithm~\cite{boruvka1926jistem} dynamically evolves during training to preserve both geometric topology and semantic coherence. 

This dynamic tree scanning-based approach effectively mitigates the two-dimensional context fragmentation problem caused by one-dimensional scanning in the traditional SSM architecture, maintains the semantic integrity and feature continuity of different parts of the underwater instances, and effectively improves the model's ability to understand the local details of the instance objects. Through patch morphing and dynamic tree pruning, the tree structure gradually focuses on instance-level feature clusters, thus accurately segmenting instance objects in complex underwater scenes.

\subsection{Hidden State Weaken (HSW)}
The R-channel of underwater images is severely degraded, and the image color is dominated by G and B. At the same time, due to the complexity and variability of the marine environment, backgrounds that are particularly similar in color and features to instances can provide significant interference for instance feature extraction. Mamba has a unique hidden state updating mechanism, and the arranged patch sequences will be reflected by the importance of the hidden state computation, providing a path to suppress the influence of background interference from another perspective.

In this regard, we design Hidden State Weaken Module (HSW) for underwater complex environments, which suppresses the influence of background patches on hidden state updates through a foreground-background separation mechanism based on Ncut~\cite{wang2022tokencut}. HSW proceeds through two steps:

\noindent \textbf{Ncut-Based Patch Categorization.} We perform the category determination of patches through a foreground-background separation mechanism based on the Ncut algorithm. Instead of re-establishing the graph connection to realize the first step of Ncut algorithm, we directly adopt the Minimum Spanning Tree $\mathcal{T}$ and the dynamic edge weights $w_{ij}$ established in the DTS to carry out the operation. This reduces the computational complexity of Ncut, maximizes the complementarity of computational mechanisms between modules, and makes graph processing more coherent.

We utilize the similarity that has been derived in the DTS and refer to Ncut's principles for patch partitioning, considering $\mathcal{T}$ as the set of all nodes $\mathcal{V}$. Subsequently, $\mathcal{V}$ is partitioned into two disjoint sets $A$, $B$, thus transforming the graph partitioning problem into a problem of minimizing the Ncut energy function:

\begin{equation}
    Ncut(A,B)=\frac{C(A,B)}{C(A,\mathcal{V})}+\frac{C(A,B)}{C(B,\mathcal{V})},
\end{equation}
where $C$ measures the degree of similarity between two sets. $C(A,B)$ is the total connection from node A to node B. $C(A,\mathcal{V})$ is the total connection from node A to all nodes in the graph. 

The divided foreground patch set A* and background patch set B* are obtained by minimizing the Ncut energy function:

\begin{equation}
(A^*, B^*) = \mathop{\arg\min}_{A,B} \left[ Ncut(A,B) \right].
\end{equation}

The binary mask is then generated via:
\begin{equation}
\text{Mask}(\mathbf{p}_i) = \mathbb{I}\left[ \mathbf{p}_i \in \mathcal{V} \right] = 
\begin{cases}
1, & \mathbf{p}_i \in A^* \\
0, & \mathbf{p}_i \in B^*,
\end{cases}
\end{equation}
where $\text{Mask}(\mathbf{p}_i)$ denotes the foreground-background classification function, which is used to determine whether the patch $\mathbf{p}_i$ belongs to the foreground or background.

\noindent \textbf{Hidden State Weaken.} After the Ncut operation, all patches have been labeled with foreground and background. We set a suppression weight $\phi$ to selectively suppress the influence of the background during the hidden state update process based on the patch category, thereby highlighting the dominant role of instance objects in the hidden state. This part of the processing can be represented as:

\begin{equation}
\phi_i = \begin{cases} 
1, & \text{Mask}(\mathbf{p}_i)=1 \\
\varphi \in (0,1), & \text{Mask}(\mathbf{p}_i)=0
\end{cases},
\end{equation}
where $\phi_i$ denotes the suppression weight for patch $i$,  $\varphi \in (0,1)$ is a hyper-parameter for background patches, and $l$ indicates the feature pyramid level in the multi-scale processing framework.

Based on the obtained weight coefficients, we suppress the weights of the edges connecting the background $\mathbf{p}$ after building the minimum spanning tree. For edges $e_{ij}$ connecting background patches to foreground nodes, we reformulate the hidden state update as:

\begin{equation}
    h_i = \sum_{j \in \mathcal{G}} \overline{B}_j \cdot \phi_i\mathbf{p_j} \prod_{k \in N_{ij}} \overline{A}_k,
\end{equation}
\begin{equation}
    y_i = C_i \mathrm{Norm}(h_i) + D \cdot \phi_i\mathbf{p_i},
\end{equation}
where $h_i$ is the hidden state at node $i$, $\mathcal{G}$ denotes the set of all nodes in the computational graph, $\overline{B}_j$ represents the input projection matrix for node $j$, $N_{ij}$ indicates the 4-connected neighborhood nodes of node $j$, $\overline{A}_k$ is the state transition matrix along edge $k$, $y_i$ is the output state at node $i$, $C_i$ denotes the output projection matrix, $\mathrm{Norm}(\cdot)$ stands for layer normalization operation, and $D$ is the skip connection weight matrix. 

\section{Experiments}

\begin{table*}[ht]
\setlength{\abovecaptionskip}{1pt}
\vspace{-4mm}
  \caption{Results on UIIS with our UIS-Mamba.}
  \label{tab:commands}
  \begin{tabular}{c|c|c|ccc|ccc|cccl}
    \toprule
    Method & Backbone & Params &  $\text{mAP}$ & $\text{AP}_{50}$ & $\text{AP}_{75}$ & $\text{AP}_{S}$ & $\text{AP}_{M}$ & $\text{AP}_{L}$ & $\text{AP}_{f}$ & $\text{AP}_{h}$ & $\text{AP}_{r}$\\
    \hline
    \texttt \textbf{Mask R-CNN}~\cite{he2017mask} & ResNet-50 & 50M &23.5 & 42.3 & 23.7 & 7.8 & 19.3 & 34.9 & 44.3 & 46.4 & 15.8\\
    \texttt \textbf{WaterMask R-CNN}~\cite{lian2023watermask} & ResNet-50 & 54M &  26.4 & 43.6 & 28.8 & 9.1 & 21.1 & 38.1 & 46.9 & 54.0 & 18.2\\
    \texttt \textbf{\textbf{UIS-Mamba(Ours)}} & \textbf{UIS-Mamba-T} & 56M &  \textbf{29.4} & \textbf{46.7} & \textbf{31.3} & \textbf{10.1} & \textbf{22.5} & \textbf{41.9} & \textbf{48.7} & \textbf{56.4} & \textbf{19.9}\\
    
    \hline
    
    \texttt \textbf{Mask R-CNN}~\cite{he2017mask} & ResNet-101 & 63M &  23.4 & 40.9 & 25.3 & 9.3 & 19.8 & 32.5 & 43.6 & 49.0 & 18.0\\
    \texttt \textbf{Mask Scoring R-CNN}~\cite{huang2019mask} & ResNet-101 & 79M & 24.6 & 41.9 & 26.5 & 8.4 & 20.0 & 34.3 & 44.2 & 52.8 & 16.0\\    
    \texttt \textbf{Cascade Mask R-CNN}~\cite{cai2018cascade} & ResNet-101 & 88M & 25.5 & 42.8 & 27.8 & 7.5 & 20.1 & 35.0 & 43.9 & 52.9 & 22.3\\
    \texttt \textbf{BMask R-CNN}~\cite{cheng2020boundary} & ResNet-101 & 66M &  22.1 & 36.2 & 24.4 & 5.8 & 17.5 & 35.0 & 40.7 & 50.0 & 17.7\\
    \texttt \textbf{Point Rend}~\cite{li2019underwater} & ResNet-101 & 63M &  25.9 & 43.4 & 27.6 & 8.2 & 20.2 & 38.6 & 43.3 & 54.1 & 20.6\\
    \texttt \textbf{$R^3$-CNN}~\cite{rossi2021recursively} & ResNet-101 & 77M &  24.9 & 40.5 & 27.8 & 9.7 & 21.4 & 33.6 & 45.4 & 52.2 & 20.2\\
    \texttt \textbf{Mask Transfiner}~\cite{ke2022mask} & ResNet-101 & 63M &  24.6 & 42.1 & 26.0 & 7.2 & 19.4 & 36.1 & 43.8 & 46.3 & 19.8\\
    \texttt \textbf{Mask2Former}~\cite{cheng2022masked} & ResNet-101 & 63M &  25.7 & 38.0 & 27.7 & 6.3 & 18.9 & 38.1 & 41.1 & 51.9 & 23.1\\
    \texttt \textbf{WaterMask R-CNN}~\cite{lian2023watermask} & ResNet-101 & 67M &  27.2 & 43.7 & 29.3 & 9.0 & 21.8 & 38.9 & 46.3 & 54.8 & 20.9\\
    \texttt \textbf{\textbf{UIS-Mamba(Ours)}} & \textbf{UIS-Mamba-S} & 76M &  \textbf{30.4} & \textbf{48.6} & \textbf{33.2} & \textbf{10.2} & \textbf{23.3} & \textbf{42.7} & \textbf{49.4} & \textbf{57.0} & \textbf{23.7}\\

    \hline

    \texttt \textbf{USIS-SAM}~\cite{lian2024diving} & ViT-H & 700M &  29.4 & 45.0 & 32.3 & 9.8 & 22.1 & 42.0 & 49.3 & 56.7 & 21.8\\
    \texttt \textbf{\textbf{UIS-Mamba(Ours)}} & \textbf{UIS-Mamba-B} & 115M &  \textbf{31.2} & \textbf{49.1} & \textbf{34.5} & \textbf{10.4} & \textbf{24.2} & \textbf{43.5} & \textbf{50.1} & \textbf{57.8} & \textbf{25.4}\\
    
    \bottomrule
  \end{tabular}
  \label{tab:uiis.comp}
\end{table*}

\subsection{Datasets and Metrics}

\noindent \textbf{Dataset.}
We evaluated our model on two benchmark datasets: the Underwater Instance Segmentation dataset (UIIS)~\cite{lian2023watermask} and the Underwater Salient Instance Segmentation dataset (USIS10K)~\cite{lian2024diving}. The UIIS dataset was divided into 3,937 training images and 691 validation images. For the USIS10K dataset, its 10,632 images were partitioned into training, validation, and test sets in a 7:1.5:1.5 ratio. Notably, USIS10K supports both category-agnostic saliency segmentation (class-independent) and category-aware saliency segmentation (class-specific) experiments. Both datasets encompass seven underwater instance categories: Fish, Reefs, Aquatic plants, Wrecks/ruins, Human divers, Robots, and Sea-floor. A key distinction lies in their focus: UIIS emphasizes multi-object, multi-instance detection and segmentation, while USIS10K prioritizes segmenting salient objects across varying scales.

\noindent \textbf{Evaluation Metrics.}
We adopt the standard mask Average Precision (AP) metrics~\cite{lin2014microsoft} for evaluation, including $\text{mAP}$, $\text{AP}_{50}$, $\text{AP}_{75}$, $\text{AP}_S$, $\text{AP}_M$, and $\text{AP}_L$ under varying Intersection over Union (IoU) thresholds. We also report the parameter counts of all compared models to comprehensively demonstrate their efficiency. Additionally, we provide class-specific $\text{mAP}$ results for key underwater application scenarios: fish ($\text{AP}_f$), human divers ($\text{AP}_h$), and wrecks/ruins ($\text{AP}_r$)~\cite{lian2023watermask}, to validate the model's applicability in critical domains such as underwater ecological conservation, human-robot interaction, and marine exploration.

\subsection{Implementation Details}
We implement UIS-Mamba using PyTorch and MMDetection~\cite{chen1906mmdetection}. Our model is trained on a NVIDIA A100 GPU with a batch size of 2, using SGD optimizer with an initial learning rate of 2.5e-3. Our three sizes of the backbone  are all based on GrootV's pre-trained weights~\cite{xiao2024grootvl} on ImageNet-1K, all detection and segmentation head settings are the same as WaterMask~\cite{lian2023watermask}, and all models are trained using a 3× learning schedule.
For the DTS module, we set the offset and scale range of patch inclusion between 0.8 and 1.2 times to prevent excessive deformation due to extreme localized information; we set the learnable parameter $\alpha$ to automatically balance the weights of spatial and semantic information in Eq. (6). 
We set the optimal $\phi$ value to 0.7 and provide the results for different values of this hyperparameter in Tab.\ref{tab:HSW}.
For the Ncut algorithm, we have the same settings as Tokencut~\cite{wang2022tokencut}.

\begin{table*}[ht]
\setlength{\abovecaptionskip}{5pt}
    \caption{Results on USIS10K with our UIS-Mamba.}
    \vspace{-2mm}
    \begin{center}
    \renewcommand{\arraystretch}{1.1}
    \setlength{\tabcolsep}{3mm}
    {\begin{tabular}{c|c|c|ccc|ccc}
    \toprule
    \multirow{2}{*}{Method}  & \multirow{2}{*}{Backbone}  & \multirow{2}{*}{Params}&\multicolumn{3}{c|}{Class-Agnostic} & \multicolumn{3}{c}{Multi-Class}\\ \cline{4-9}
     &  &   &  $\text{mAP}$ & $\text{AP}_{50}$ & $\text{AP}_{75}$ & $\text{mAP}$ & $\text{AP}_{50}$ & $\text{AP}_{75}$\\
    \hline
    S4Net \cite{Fan_2019_CVPR} & ResNet-50 & 47M &  32.8 & 64.1 & 27.3 & 23.9 & 43.5 & 24.4 \\
    RDPNet \cite{RDPnet_2021_TIP}  & ResNet-50 & 49M & 53.8 & 77.8 & 61.9 & 37.9 & 55.3 & 42.7\\
    OQTR \cite{OQTR_2022_TOM}  & ResNet-50 & 50M &  56.6 & 79.3 & 62.6 & 19.7 & 30.6 & 21.9\\
    WaterMask \cite{lian2023watermask} & ResNet-50 & 54M & 58.3 & 80.2 & 66.5 & 37.7 & 54.0 & 42.5 \\

    \textbf{UIS-Mamba(Ours)}  & \textbf{UIS-Mamba-T} & 56M & \textbf{62.2} & \textbf{84.0} & \textbf{71.3} & \textbf{42.1} & \textbf{59.6} & \textbf{48.3}\\
    \hline
    
    RDPNet \cite{RDPnet_2021_TIP} & ResNet-101  & 66M  & 54.7 & 78.3 & 63.0 & 39.3 & 55.9 & 45.4\\
    WaterMask \cite{lian2023watermask}& ResNet-101   & 67M & 59.0 & 80.6 & 67.2 & 38.7 & 54.9 & 43.2\\

    \textbf{UIS-Mamba(Ours)} & \textbf{UIS-Mamba-S} & 76M & \textbf{63.1} & \textbf{85.1} & \textbf{72.0}& \textbf{44.5} & \textbf{61.5} & \textbf{51.1}\\
    \hline
    
    SAM+BBox \cite{SAM_2023_ICCV} & ViT-H & 641M  & 45.9 & 65.9 & 52.1 & 26.4 & 38.9 & 29.0 \\
    SAM+Mask \cite{SAM_2023_ICCV} & ViT-H & 641M  & 55.1 & 80.2 & 62.8 & 38.5 & 56.3 & 44.0 \\
    RSPrompter \cite{chen2023rsprompter}  & ViT-H & 632M & 58.2 & 79.9 & 65.9 & 40.2 & 55.3 & 44.8\\
    USIS-SAM \cite{lian2024diving} & ViT-H & 701M & 59.7 & 81.6 & 67.7 & 43.1 & 59.0 & 48.5\\
 
    \textbf{UIS-Mamba(Ours)} & \textbf{UIS-Mamba-B} & 115M & \textbf{63.8} & \textbf{86.0} & \textbf{72.8}& \textbf{46.2} & \textbf{63.2} & \textbf{53.4}\\

    \bottomrule
    \end{tabular}}
    \end{center}
    \label{tab:usis.comp}
\end{table*}

\subsection{Main Results}
\noindent \textbf{Instance Segmentation.}
As shown in Tab.\ref{tab:uiis.comp}, we compare our method with the current state-of-the-art methods on the UIIS~\cite{lian2023watermask} dataset. For UIS-Mamba-T backbone, our method achieves 29.4, 46.7, and 31.3 AP in mAP, $\text{AP}_{50}$, and $\text{AP}_{75}$. It is 3.0, 3.1, and 2.5 AP in mAP, $\text{AP}_{50}$, and $\text{AP}_{75}$ higher than SOTA method WaterMask R-CNN with ResNet-50 backbone~\cite{lian2023watermask}. In addition, for $\text{AP}_S$, $\text{AP}_M$, $\text{AP}_L$ in terms of instance size and $\text{AP}_f$, $\text{AP}_h$, $\text{AP}_r$ in terms of instance class, our method also achieves comprehensive improvement and shows the performance of dealing with various kinds of underwater instance objects. We compare our approach on the UIS-Mamba-S backbone with more SOTA approaches, including those using the Transformer architecture. Compared with the state-of-the-art Mask2Former~\cite{Cheng2022Mask2Former}, Point Rend~\cite{kirillov2020pointrend},WaterMask R-CNN with ResNet-101 backbone~\cite{lian2023watermask}, our method achieves 30.4 mAP, surpassing 4.7, 4.5, and 3.2 AP, respectively, while realizing all other metrics. Further scaling backbone, UIS-Mamba-B reaches 31.2 mAP, surpassing USIS-SAM with ViT-H~\cite{lian2024diving} by 1.8mAP, but the number of parametrics is only 16.4\% of its.

\noindent \textbf{Class-Agnostic Salient Instance Segmentation.}
Class-agnostic salient instance segmentation can be essentially understood as foreground instance segmentation exclusively focusing on salient regions in images. Consequently, the model requires  robust capability to localize salient areas, and the segmented salient instance masks must strictly align with instance objects to achieve favorable experimental results. Furthermore, due to the complexity and substantial diversity of underwater images, a single image may contain multiple salient instances, posing additional challenges for the model. As shown in Tab.~\ref{tab:usis.comp}, we extensively compare with the state-of-the-art (SOTA) methods on the USIS10K dataset~\cite{lian2024diving}. For the UIS-Mamba-T backbone, our method achieves 62.2, 84.0, and 71.3 AP in mAP, $\text{AP}_{50}$, and $\text{AP}_{75}$, surpassing the SOTA method WaterMask R-CNN~\cite{lian2023watermask} with ResNet-50 backbone by 3.9, 3.8, and 4.8 AP in mAP, $\text{AP}_{50}$, and $\text{AP}_{75}$ respectively. It also outperforms other advanced methods RPDNet~\cite{RDPnet_2021_TIP} and OQTR~\cite{OQTR_2022_TOM} with ResNet-50 backbone by 8.4 and 5.6 AP in mAP. When rigorously 
comparing with SOTA methods using the UIS-Mamba-S backbone, including RDPNet~\cite{RDPnet_2021_TIP} and WaterMask R-CNN~\cite{lian2023watermask} with ResNet-101 backbone, our method achieves 63.1, 85.1, and 72.0 AP in mAP, $\text{AP}_{50}$, and $\text{AP}_{75}$, surpassing them by 8.4 and 4.1 AP in mAP respectively. By further scaling the backbone, UIS-Mamba-B ultimately attains 63.8, 86.0, and 72.8 AP in mAP, $\text{AP}_{50}$, and $\text{AP}_{75}$, outperforming transformer-based methods including SAM+Mask~\cite{SAM_2023_ICCV}, RSPrompter~\cite{chen2023rsprompter}, and USIS-SAM~\cite{lian2024diving} with ViT-H by 8.7, 5.6, and 4.1 AP in mAP respectively, conclusively demonstrating the superior performance of our model.

\noindent \textbf{Multi-Class Salient Instance Segmentation.}
Multi-class salient instance segmentation can be conceptually considered as a fusion of tasks from class-independent salient instance segmentation and salient instance class prediction. Meanwhile, due to the diversity of underwater instance objects, the morphological features exhibited by similar objects are more complex and varied than those of the ground environment, which makes the model fitting and understanding extremely difficult, and requires a more targeted design to effectively adapt to the underwater images. As shown in Tab.~\ref{tab:usis.comp}, we comprehensively compare our method with the current state-of-the-art methods on the USIS10K dataset~\cite{lian2024diving}, and UIS-Mamba continues to hold the advantage with accurate foreground instance capture and 
background hidden state weakening in the case of multiple categories. For UIS-Mamba-T backbone, our method achieves 42.1, 59.6, and 48.3 AP in mAP, $\text{AP}_{50}$, and $\text{AP}_{75}$, respectively, surpassing the state-of-the-art WaterMask R-CNN~\cite{lian2023watermask} with ResNet-50 backbone by 4.4, 5.6, and 5.8 AP in mAP, $\text{AP}_{50}$, and $\text{AP}_{75}$. Additionally, our method outperforms other leading methods, including RDPNet~\cite{RDPnet_2021_TIP} and OQTR~\cite{OQTR_2022_TOM}, by 22.4 and 4.4 AP in mAP, respectively. We compared with SOTA methods on UIS-Mamba-S backbone including RDPNet~\cite{RDPnet_2021_TIP} and WaterMask R-CNN~\cite{lian2023watermask} with ResNet-101 backbone, our method achieves 44.5, 61.5, and 51.1 AP in mAP, $\text{AP}_{50}$, and $\text{AP}_{75}$, surpassing 5.2, 5.8 AP in mAP, respectively. Further scaling backbone, UIS-Mamba-B achieves 46.2, 63.2, and 53.4 AP in mAP, $\text{AP}_{50}$, and $\text{AP}_{75}$, respectively. Compared to the SAM-based method , our method
surpasses SAM+Mask~\cite{SAM_2023_ICCV}, RSPrompter~\cite{chen2023rsprompter}, USIS-SAM~\cite{lian2024diving} with ViT-H 7.7, 6.0, and 3.1 AP in mAP, respectively. The advantage of our method is more obvious in multi-category significant instance segmentation.

\begin{figure*}[htbp]
	\center{\includegraphics[width=18cm]{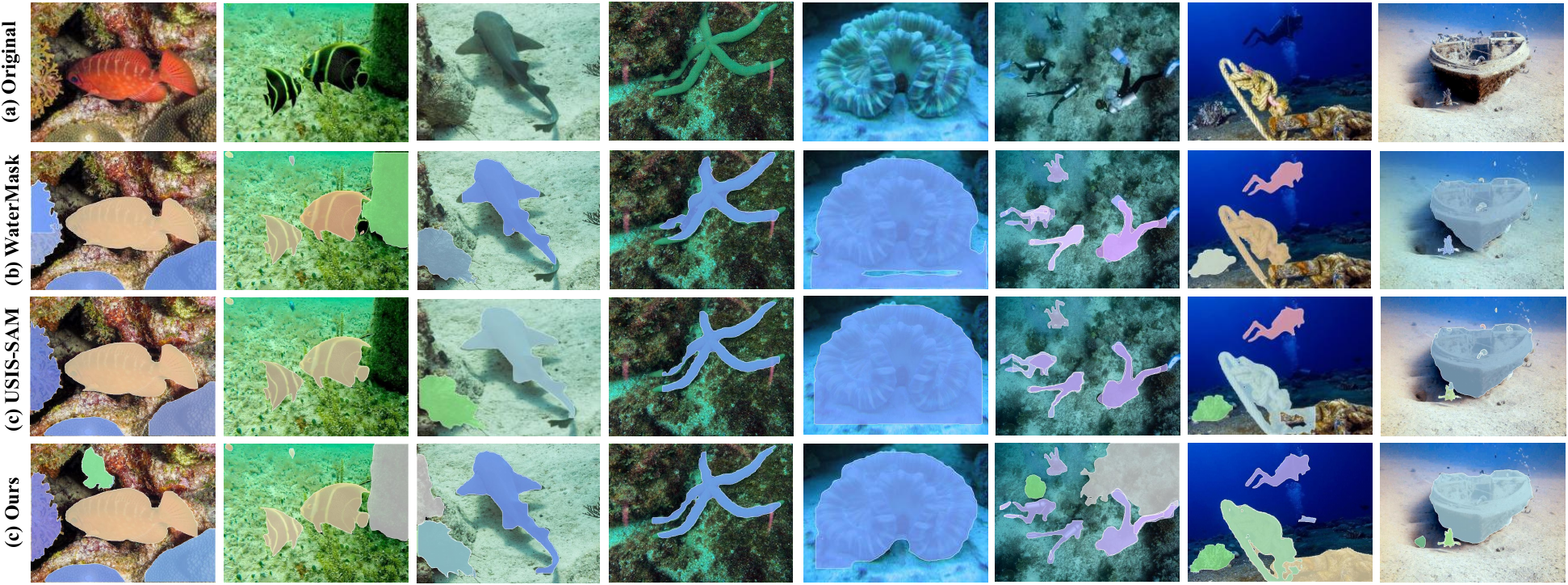}} 
	\caption{Qualitative comparison on the UIIS and USIS10K dataset. Each class of instance in the same image is represented by a unique color, and the segmented mask is superimposed on the image.}
	\label{fig:uiis}
\end{figure*}

\subsection{Ablation Study}
In this subsection, we conduct four ablation studies on UIIS dataset with UIS-Mamba-T to systematically validate: (1) the effectiveness of individual modules, (2) internal configuration analysis of DTS Module, (3) the optimal configuration in HSW Module, and (4) performance comparisons of different instance segmentation heads.

\begin{table}[th]
\vspace{-3mm}
 \setlength{\abovecaptionskip}{1pt} 
  \caption{Ablation study of our contributions.}
  \label{tab:module}
  \begin{tabular}{cccccccl}
    \toprule
    Mamba & DTS & HSW & Params & $mAP$ & $AP_{50}$ & $AP_{75}$ \\
    \midrule
              &           &           & 54M & 26.4 & 43.6 & 28.8\\
    \ding{52} &           &           & 52M & 27.1 & 45.1 & 29.3\\
    \ding{52} & \ding{52} &           & 55M & 28.9 & 46.2 & 30.9\\
    \ding{52} &           & \ding{52} & 53M & 28.2 & 46.1 & 30.2\\
    \ding{52} & \ding{52} & \ding{52} & \textbf{56M} & \textbf{29.4} & \textbf{46.7} & \textbf{31.3}\\
  \bottomrule
\end{tabular}
\setlength{\belowcaptionskip}{1pt}
\vspace{-3mm}
\label{tab:Module}
\end{table}
\noindent \textbf{Effectiveness of Individual Modules.} We comprehensively show the results of the following ablation experiments in Tab.\ref{tab:Module}: (1) whether to use Mamba backbone instead of ResNet-50, (2) whether to use the DTS module, and (3) whether to use the HSW module. The experiments show that after making all three improvements, the model has a positive metric improvement while keeping the number of parameters low. The baseline on the first line is WaterMask R-CNN~\cite{lian2023watermask} with ResNet-50 backbone. Replacing ResNet-50 with Mamba backbone~\cite{xiao2024grootvl} improved the metrics by 0.7 AP in mAP while maintaining the original number of parameters; after using the DTS module and the HSW module separately, the metrics improved by 1.8, 1.1 AP in mAP, respectively; and after using the DTS together with the HSW, the metrics improved to 29.4 AP in mAP. Notably, the model achieves the optimal performance with a modest increase of only 2M parameter quantity.

\begin{table}[th]
\vspace{-3mm}
 \setlength{\abovecaptionskip}{1pt} 
  \caption{Ablation study of DTS module.}
  \label{tab:dts}
  \begin{tabular}{cccccl}
    \toprule
    Offsets & Scales & Weights & $mAP$ & $AP_{50}$ & $AP_{75}$ \\
    \midrule
              &           &           & 27.1 & 45.1 & 29.3\\
    \ding{52} &           &           & 27.7 & 45.6 & 29.7\\
              & \ding{52} &           & 27.5 & 45.4 & 29.5\\
    \ding{52} & \ding{52} &           & 28.4 & 45.8 & 30.1\\
    \ding{52} & \ding{52} & \ding{52} & \textbf{28.9} & \textbf{46.2} & \textbf{30.9}\\
  \bottomrule
\end{tabular}
\setlength{\belowcaptionskip}{1pt}
\vspace{-3mm}
\label{tab:DTS}
\end{table}
\noindent \textbf{Ablation Study on the DTS Module.} 
We present the results of three targeted improvements in the DTS module in Tab.~\ref{tab:DTS}. 
The offset is predicted to shift patches toward more critical regions, while the scaling operation is designed to better preserve local structures. 
The weight definition based on these improvements, combined with the minimum spanning tree (MST), enables more effective fusion of spatial and semantic information. 
All components contribute to performance enhancement. 
As decoupled in Tab.~\ref{tab:DTS}, applying only the offset or scaling operation improves mAP by 0.7 and 0.4 AP, respectively. 
When incorporating distance information into the calculation of edge weights, the mAP increases from 28.4 to 28.9 AP, demonstrating the model's improved capability to capture both global and local features of underwater instances.

\begin{table}[t]
\vspace{-3mm}
 \setlength{\abovecaptionskip}{1pt} 
  \caption{Impact of the hyperparameter $\varphi$.}
  \begin{tabular}{ccccl}
    \toprule
    $\varphi$ & $mAP$ & $AP_{50}$ & $AP_{75}$ \\
    \midrule
    0   & 27.1 & 45.1 & 29.3\\
    0.5 & 27.7 & 45.5 & 29.6\\
    0.6 & 28.0 & 45.6 & 29.9\\
    \textbf{0.7} & \textbf{28.2} & \textbf{46.1} & \textbf{30.2}\\
    0.8 & 27.9 & 45.7 & 29.7\\
  \bottomrule
\end{tabular}
\setlength{\belowcaptionskip}{1pt}
\vspace{-3mm}
\label{tab:HSW}
\end{table}
\noindent \textbf{Ablation Study on the HSW Module.} As shown in Tab.\ref{tab:HSW}, we analyze the impact of the hyperparameter $\varphi$ in Equation~(11) on the HSW module. Experiments with $\varphi$ values adjusted in 0.1 increments reveal that $\varphi = 0.7$ achieves optimal performance. This demonstrates that appropriately suppressing background hidden state updates (controlled by $\varphi$) enhances segmentation accuracy by focusing information flow on target instances. Notably, the optimal $\varphi < 1.0$ implies retaining partial background information improves model robustness, due to preserved environmental cues critical for underwater instance reasoning.

\begin{table}[H]
 \setlength{\abovecaptionskip}{1pt} %
  \caption{Impact of different instance segmentation heads.}
  \label{tab:mask}
  \begin{tabular}{cccccccl}
    \toprule
    Head & $mAP$ & $AP_{50}$ & $AP_{75}$ & $AP_{S}$ & $AP_{M}$ & $AP_{L}$ \\
    \midrule
    MaskRCNN & 28.3 & 45.6 & 30.7 & 9.6 & 21.1 & 39.3 \\
    \textbf{WaterMask} & \textbf{29.4} & \textbf{46.7} & \textbf{31.3} & \textbf{10.1} & \textbf{22.5} & \textbf{41.9} \\
  \bottomrule
\end{tabular}
\setlength{\belowcaptionskip}{1pt}
\label{tab:WaterMask}
\end{table}

\noindent \textbf{Comparative Analysis of different instance segmentation heads.} Tab.~\ref{tab:WaterMask} demonstrates the performance impact when integrating the UIS-Mamba backbone with either Mask R-CNN~\cite{he2017mask} or the underwater-optimized WaterMask~\cite{lian2023watermask} detection/segmentation heads. The results reveal that WaterMask enables superior underwater feature comprehension, particularly enhancing segmentation accuracy for small instances and complex-boundary objects.

\subsection{Qualitative Results}
We also show some qualitative visual comparisons with WaterMask and USIS-SAM on the UIIS and USIS10K test set in Fig.\ref{fig:uiis}. 
This shows that our approach consistently succeeds in correctly categorizing individual instances and does a good job of capturing well-hidden instances, such as Aquatic plants in the first column of the figure, Reefs in the second column, and Wrecks in the seventh column. The model also does a good job of segmenting instances as well as the overall shape of salient instances, even in challenging regions, such as those shown in the fifth and eighth columns of the figure. Meanwhile, our method produces more accurate boundaries and details, and identifies well to specific sites, such as Fish in the second and third columns, and the Starfish in the fourth column in Fig.\ref{fig:uiis}.

\section{Conclusions}
In this paper, we propose the first Mamba-based underwater instance segmentation model, UIS-Mamba, to port Mamba to underwater tasks through two improved modules to address the problems of segmentation confusion and semantic ambiguity in challenging underwater scenes. The DTS module maintains the continuity of the internal features of the instance objects by allowing the graph structure to be dynamically shifted and scaled to guide the minimum spanning tree and provides dynamic local sense fields. The HSW module suppresses interference from complex backgrounds and focuses the information flow of state propagation to the instances themselves through a hidden state weakening mechanism based on Ncut. UIS-Mamba achieves state-of-the-art performance on both the UIIS and the USIS10K datasets, while keeping the number of parameters and computational complexity low.

\begin{acks}
This work was supported in part by the Taishan Scholar Project of Shandong Province under Grant tsqn202306079, in part by the National Natural Science Foundation of China Grant 62471278, and in part by the Research Grants Council of the Hong Kong Special Administrative Region, China under Grant STG5/E-103/24-R.
\end{acks}

\bibliographystyle{ACM-Reference-Format}
\balance
\bibliography{main}

\end{document}